\documentclass[letterpaper,10pt,conference]{ieeeconf}

\usepackage{amsmath}
\usepackage{amsfonts}
\usepackage{amssymb}

\usepackage{amsthm}
\usepackage[ruled,vlined]{algorithm2e}
\usepackage{mathtools}
\usepackage{tabularx}
\usepackage{graphicx}
\usepackage{subfigure}
\usepackage{enumerate}
\usepackage{float}
\usepackage{url}
\usepackage{verbatim}
\usepackage{cite}
\usepackage[linkcolor=black,citecolor=black,urlcolor=black,colorlinks=true]{hyperref}
\usepackage{empheq}
\usepackage{siunitx}

\graphicspath{{figures/}}
\IEEEoverridecommandlockouts
\DeclareMathOperator*{\vol}{vol}
\DeclareMathOperator*{\diag}{diag}
\newcommand{\tp}{^{\mathrm{T}}}

\newcommand{\df}[1]{\mathrm{d}{#1}}
\newcommand{\rbrac}[1]{({#1})}
\newcommand{\rBrac}[1]{\left({#1}\right)}

\newcommand{\cbrac}[1]{\{{#1}\}}
\newcommand{\cBrac}[1]{\left\{{#1}\right\}}

\newcommand{\sBrac}[1]{\left[{#1}\right]}

\newcommand{\norm}[1]{\Vert{#1}\Vert}
\newcommand{\Norm}[1]{\left\Vert{#1}\right\Vert}

\begin{document}
    \title{\LARGE\bf Robust Trajectory Planning for Spatial-Temporal \\ Multi-Drone Coordination in Large Scenes}
    \author{Zhepei Wang, Chao Xu, and Fei Gao
        \thanks{All authors are with both the College of Control Science and Engineering, Zhejiang University, Hangzhou 310027, China, and the Huzhou Institute, Zhejiang University, Huzhou 313000, China. {\tt\{wangzhepei, cxu, fgaoaa\}@zju.edu.cn}}
    }

    \maketitle
    \thispagestyle{empty}
    \pagestyle{empty}

\begin{abstract}
    In this paper, we describe a robust multi-drone planning framework for high-speed trajectories in large scenes. It uses a free-space-oriented map to free the optimization from cumbersome environment data. A capsule-like safety constraint is designed to avoid reciprocal collisions when vehicles deviate from their nominal flight progress under disturbance. We further show the minimum-singularity differential flatness of our drone dynamics with nonlinear drag effects involved. Leveraging the flatness map, trajectory optimization is efficiently conducted on the flat outputs while still subject to physical limits considering drag forces at high speeds. The robustness and effectiveness of our framework are both validated in large-scale simulations. It can compute collision-free trajectories satisfying high-fidelity vehicle constraints for hundreds of drones in a few minutes.
\end{abstract}

\section{Introduction}
\label{sec:Introduction}
Multi-drone coordination is receiving increasing attention as a fundamental problem in various applications such as urban delivery, exploration, and inspection. It often requires robust planning for concurrent long-distance flights in vast spaces. The coordination should also be tolerant of realistic factors from both the environment and vehicles. The high problem dimension and the huge environment data further prevent existing algorithms from being applicable to large scenes. Most of them consider over-simplified safety criteria and system dynamics in relatively short-range flights.

Several practical problems exist in multi-drone planning. Firstly, planning algorithms frequently need accessing map data for obstacle information~\cite{Bialkowski2016EfficientCC}. In large scenes, an obstacle-oriented map can become quite cumbersome, making itself a computational bottleneck. Secondly, robust planning for multi-drone should exploit the flexibility in both space and time aspects while the latter is often ignored in the literature. Thirdly, unexpected disturbances can make vehicles deviate a lot from nominal trajectories. In this case, potential reciprocal collisions pose threats to the whole system. Therefore, mismatches between actual flight progress and the planned one easily invalidate distance-based safety criteria ensured in the planning phase. Fourthly, high-speed flights admittedly improve task efficiency in large scenes. However, aerodynamic drag effects cannot be ignored when ensuring the physical limits of vehicles. These also make oversimplified feasibility criteria insufficient here.

In this paper, we propose a robust framework for multi-drone planning in large scenes. The framework is built upon three criteria for flight coordination. An obstacle avoidance criterion ensures that no collision occurs between vehicles and environments. We use a free-space-oriented map instead where a union of polyhedra tightly approximates all free configurations. The MINCO~\cite{Wang2021Gcopter} trajectory is then adopted for online spatial-temporal optimization within polyhedron-shaped corridors. We also design a reciprocal safety criterion via a space-time ``capsule" constraint. It makes large flight errors tolerable in time and position. Therefore, our planning results are robust against reasonable perturbations. To ensure a dynamic feasibility criterion, we show the differential flatness for our drones subject to nonlinear drags. The flatness map makes it possible to enforce user-defined physical limits via penalty functionals supported by MINCO.

Summarizing our contributions in this work:
\begin{itemize}
    \item {A map polyhedronization scheme with corridor generation is proposed for online free space query;}
    \item {A space-time capsule constraint is designed for robust reciprocal safety against large flight errors;}
    \item {Minimum-singularity differential flatness is shown for our drones subject to nonlinear drag effects;}
    \item {A systematic way to robust trajectory planning is provided for multi-drone coordination with physical limits.}
\end{itemize}

\begin{figure}[t]
    \centering
    \includegraphics[width=0.95\columnwidth]{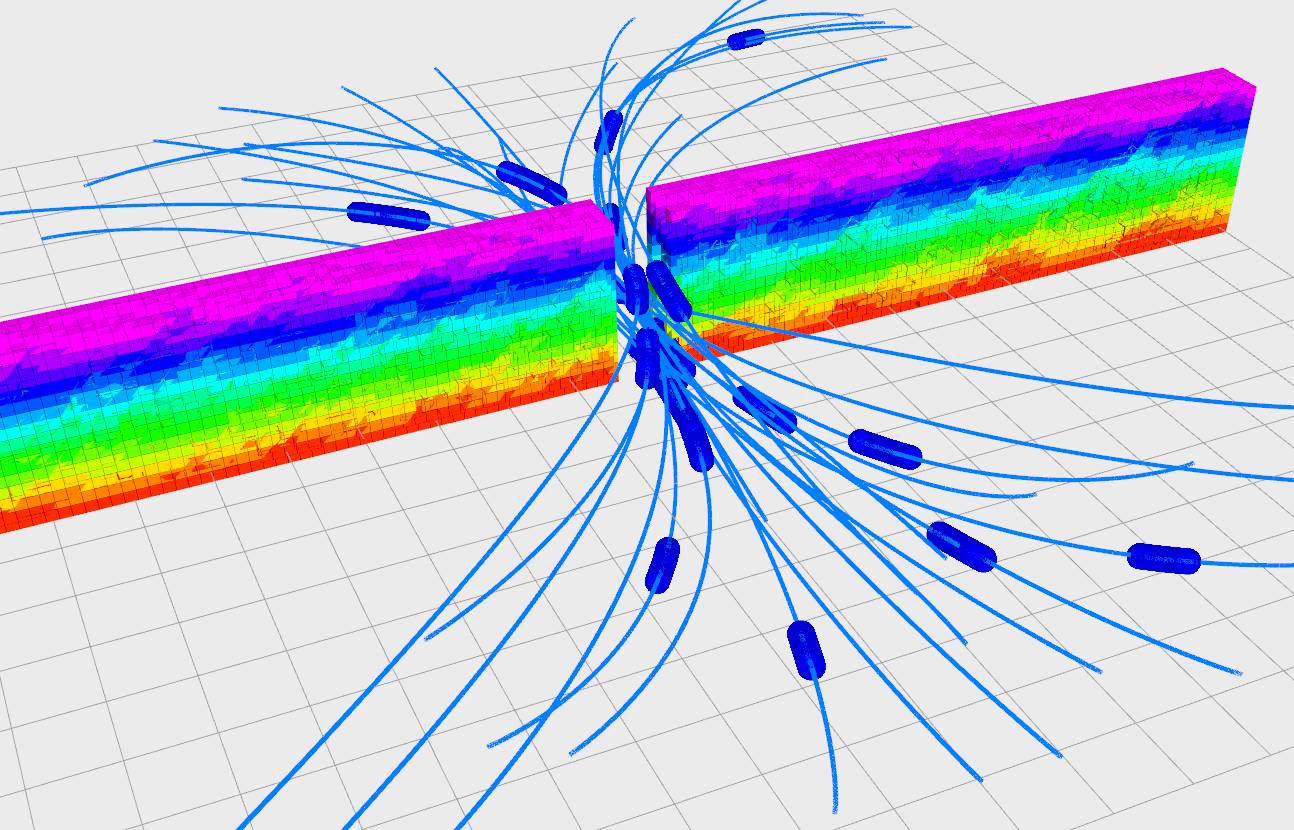}
    \caption{These are $20$ drones flying through a narrow gap in opposite directions concurrently. The blue ``capsules" indicate the space-time uncertainty of vehicles. Each vehicle can deviate from the timestamp and position of its nominal trajectory. Our robustness is guaranteed by collision avoidance between capsules via spatial-temporal planning.\label{fig:RobustnessTestViz}}
    \vspace{-0.5cm}
\end{figure}

\section{Related Work}
Planning for multi-drone coordination is inherently a high-dimensional problem even if safety is the only requirement. To reduce difficulty, reciprocal collision avoidance~\cite{Van2011ReciprocalCA} computes the feasible velocity in a decentralized way. Conflict-based search~\cite{Sharon2015ConflictBS} conducts centralized graph search instead, with complexity dominated by the number of conflicts rather than agents. Their nonsmooth results do not suit high-order dynamics. Some methods formulate the joint planning into a Mixed Integer Quadratic Program (MIQP)~\cite{Mellinger2012MixedQP} or a Sequential Convex Program (SCP)~\cite{Augugliaro2012GenerationSCP} to generate smooth trajectories in a centralized way, while they are not easily scaled to a large vehicle number. In~\cite{Tang2016SafeCTG}, separating hyperplanes are applied to convex hulls of trajectories, forming a decoupled QP for each individual in every refinement iteration. A drawback is that obstacle-free environments are assumed. In~\cite{Honig2018TrajectorySWARM}, separating planes also form the safe flight corridor for a vehicle, which excludes all other dynamic and static obstacles. The planning is done by iterative refinement based on guaranteed safe results from~\cite{Sharon2015ConflictBS}. To reduce conservativeness, these hyperplanes are treated as decision variables of a Nonlinear Program (NLP) for planning~\cite{Tordesillas2020MaderTPMA}. However, all these methods do not consider real physical limits and temporal optimization. To assure higher-fidelity dynamics, the learned aerodynamic
interactions are incorporated into full quadrotor dynamics~\cite{Shi2021NeuralSW}, showing the high stability in close-proximity coordination flights. Optimization-based temporal scheduling is performed in~\cite{William2019TemporalSOMP} while it incorporates integer variables and assumes obstacle-free environments. Different from existing work, our work accomplishes spatial-temporal coordination for multi-drone in an incremental way. It further handles state-input limits and nonlinear drag effects during high-speed flights.

\section{Preliminaries}
\subsection{Map Polyhedronization}
Safe flight corridors are convenient to encode free-space information into trajectory planning while conquering nonsmoothness in discrete environment data. Instead of online construction, we propose map polyhedronization as a pre-processing step for corridor-based trajectory planning. The process is done in advance for any large-scale fixed scene, such as occupancy grids, point clouds, or triangle meshes of digital maps.

Let $\mathcal{F}$ denote the free configuration in a given map. By map polyhedronization we mean finding a set of polyhedra $\mathcal{P}_i\subseteq\mathcal{F}$ such that their union approximates $\mathcal{F}$ at a satisfactory filling rate, i.e.,
\begin{equation}
\vol\rBrac{\bigcup_{i=1}^{M_\mathcal{P}}\mathcal{P}_i}>\rBrac{1-\epsilon}\vol\rBrac{\mathcal{F}},
\end{equation}
where $\vol\rBrac{\cdot}$ means the volume and
\begin{equation}
\mathcal{P}_i=\cBrac{x\in\mathbb{R}^3~\Big|~A_ix\preceq{b_i}}.
\end{equation}

\begin{figure}[ht]
    \centering
    \includegraphics[width=1.0\columnwidth]{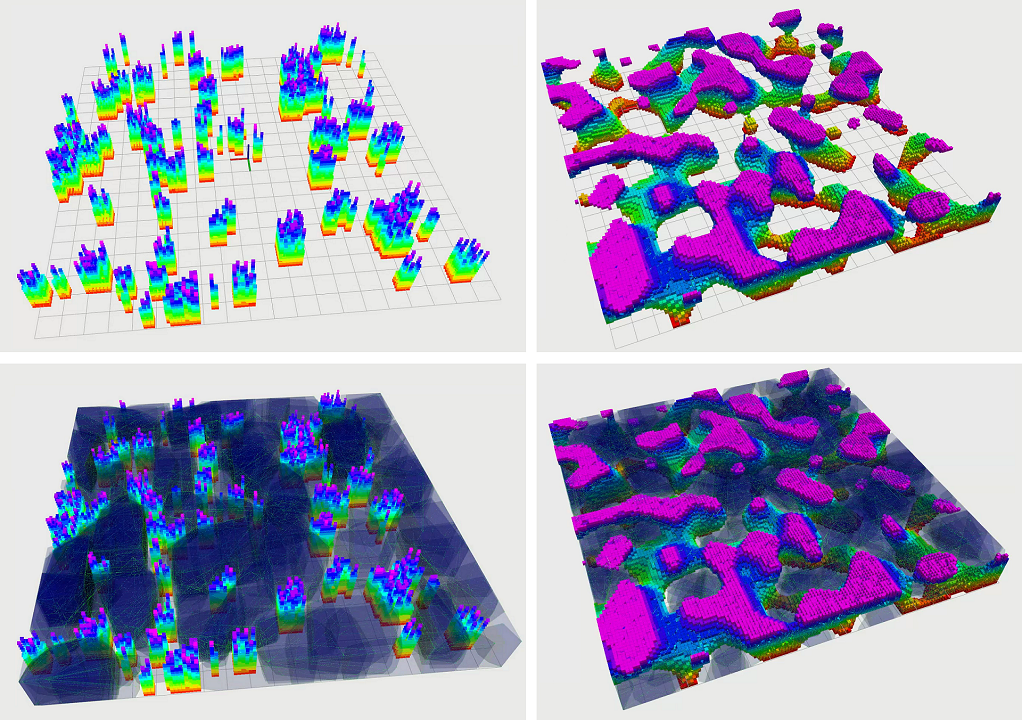}
    \caption{The top figures show two kinds of obstacle environments. The bottom figures show polyhedronization of these maps. All free spaces are tightly filled by unions of convex polyhedra, colored in dark blue.\label{fig:PolyhedronizedMaps}}
    \vspace{-0.5cm}
\end{figure}

There are many algorithms\cite{Deits2015ComputingIRIS,Liu2017PlanningDF,Gao2020TeachRepeatReplanAC,Zhong2020GeneratingLCP} able to generate a free polyhedron wrapping a given seed. To accomplish the map polyhedronization, we first select a random position $x\in\mathcal{F}\setminus\bigcup_{i=1}^{j}\mathcal{P}_i$ by rejection sampling. Any algorithm mentioned is applied to generate a $\mathcal{P}_{j+1}$. This procedure repeats for $j\leftarrow{j+1}$ until its rejection rate exceeds $1-\epsilon$. Two examples are also provided in Figure~\ref{fig:PolyhedronizedMaps}. Axis-aligned bounding boxes of these polyhedra are managed by a multi-level segment tree for three-dimensional
stabbing queries~\cite{De1997ComputationalG}. Consequently, for any position $x$ in the covered free space, it becomes efficient to obtain an outer polyhedron, denoted by $\mathcal{P}(x)$.

\subsection{Problem Statement}
Consider a set of drones conducting concurrent flights in one scene $\mathcal{F}$. The $i$-th vehicle's position is $r_i(t):\mathbb{R}\mapsto\mathbb{R}^3$ at an absolute timestamp $t$. A new flight mission from $p_o$ to $p_f$ is to start at $t_o$, while its trajectory $r(t)$ should not conflict with early existed missions. Therefore, it must follow some criteria for obstacle avoidance, reciprocal safety, and dynamic feasibility.

The obstacle avoidance criterion requires the vehicle to fly in the covered free configuration $\tilde{\mathcal{F}}=\bigcup_{i=1}^{M_\mathcal{P}}\mathcal{P}_i$, i.e.,
\begin{equation}
\label{eq:ObstacleAvoidanceCriterion}
r(t)\in\tilde{\mathcal{F}},~\forall t\in[t_o,t_f],
\end{equation}
where $T_\Sigma\in\mathbb{R}_{>0}$ is the total duration and $t_f=t_o+T_\Sigma$. The reciprocal safety criterion requires the robustness against unexpected disturbance. Thus we consider the space-time capsule constraint, ensuring safety margins in space and time aspects are both considered. Specifically,
\begin{subequations}
    \label{eq:ReciprocalSafetyCriterion}
    \begin{align}
    &\Norm{r(\alpha)-r_i(\beta)}_W\geq2M_r,\\
    &~\forall\alpha\in[t-M_d,t+M_d]\cap[t_o,t_f],\\
    &~\forall\beta\in[t-M_d,t+M_d],\\
    &~\forall t\in[t_o,t_f],
    \end{align}
\end{subequations}
where $W=\diag\cbrac{1,1,w}$ enlarges the margin vertically to prevent downwash interference if $w<1$. The scalars $M_r$ and $M_d$ are spatial and temporal margins for individuals, respectively. The dynamic feasibility criterion requires vehicle states and inputs such as the collective thrust and body rate to fulfill physical limits when the drone tracks $r(t)$ even with significant air drag at a high speed.

\section{Method}
\subsection{Just-In-Time Corridor Generation}
Safe flight corridors are essentially sequences of polyhedra. Corridors ensure collision-free flights by excluding static obstacles while keeping parameters as compact as possible. Thus, it is well-suited for large-scale maps with redundant data for navigation. We conduct just-in-time corridor generation for a newly occurred flight mission based on the interface $\mathcal{P}(x)$ of a previously polyhedronized map.

Firstly, we apply the Informed RRT*~\cite{Gammell2018InformedRRT} to obtain an approximately shortest path from $p_o$ to $p_f$ in $\tilde{\mathcal{F}}$. Our main concern here is that in large-scale scenes, energy consumption is more relevant to the distance covered. Denote by $\bar{r}(l):[0,L]\mapsto\mathbb{R}^3$ a arc-length-parameterized path. Based on the collision-free path, the generation of a homotopic safe flight corridor $\mathcal{S}$ is given by the algorithm below, also visualized in Figure~\ref{fig:ShortestPath}.

\begin{algorithm}
    \caption{Homotopic Corridor Generation}
    \label{alg:HomotopicCorridorGeneration}
    \KwIn{Path $\bar{r}(l)$ and Polyhedronized Map $\mathcal{P}(x)$}
    \KwOut{Safe Flight Corridor $\mathcal{S}$}
    \Begin
    {
        $\mathcal{S}\leftarrow\cBrac{},~l\leftarrow0$\;
        \While{$l\leq{L}$}
        {
            $x\leftarrow{\bar{r}(l)}$\;
            $\bar{\mathcal{P}}\leftarrow\mathcal{P}(x)$\;
            $\mathcal{S}.\textbf{append}\rBrac{\bar{\mathcal{P}}}$\;
            $l \leftarrow \max_{\theta\in[l,L]}{\theta},~s.t.~\bar{r}(\vartheta)\in\bar{\mathcal{P}},~\forall\vartheta\in[l,\theta]$\;
        }
        \Return{$\mathcal{S}$};
    }
\end{algorithm}

Algorithm~\ref{alg:HomotopicCorridorGeneration} actually finds a convex cover of the guiding path as soon as a new flight mission occurs. All polyhedra have been precomputed and can be online accessed directly. Thus the corridor construction is just-in-time and efficient. $\mathcal{S}$ provides a large room for consequent trajectory planning subject to various constraints. For convenience, we denote $\mathcal{S}$ by a polyhedron sequence $\cBrac{\mathcal{P}_1,\mathcal{P}_2,\dots,\mathcal{P}_M}$ hereafter.

\subsection{Spatial-Temporal Planning in Flight Corridors}
\label{sec:SpatialTemporalPlanningInCorridors}

\begin{figure}[t]
    \centering
    \includegraphics[width=0.95\columnwidth]{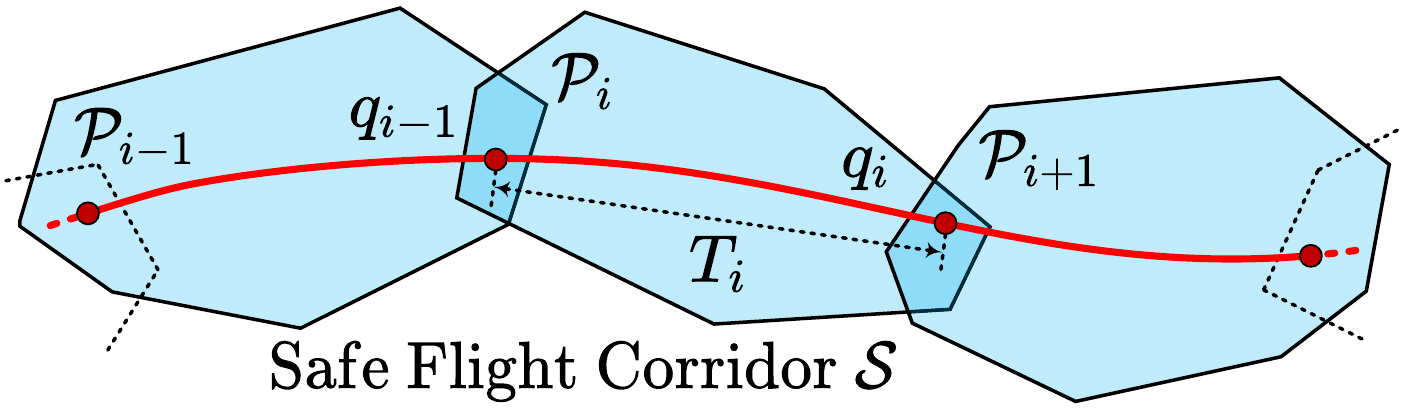}
    \caption{Every waypoint $q_i$ is sequentially assigned in to the intersection of consecutive polyhedra $\mathcal{P}_i\cap\mathcal{P}_{i+1}$. This assignment forms a convex constraint for each piece, instead of the nonconvex constraint $\mathcal{S}$. The trajectory is only parameterized by waypoints and durations $T_i$.\label{fig:TrajectoryInSFC}}
    \vspace{-0.5cm}
\end{figure}

For efficiency, we conduct optimization in the flat-output space of multicopters such that all differential constraints from dynamics are fulfilled by default. We adopt the MINCO representation~\cite{Wang2021Gcopter} to conduct spatial-temporal deformation of the flat-output trajectory.

An $s$-order MINCO trajectory is indeed a $2s$-order polynomial spline with constant boundary conditions. It provides a linear-complexity smooth map from intermediate points $q$ and a time allocation $T$ to the coefficients of splines, which is denoted as $\mathcal{M}(q,T):\mathbb{R}^{3\times(M-1)}\times\mathbb{R}^{M}_{>0}\mapsto\mathbb{R}^{2Ms\times3}$. A spline with $c=\mathcal{M}(q,T)$ is exactly the unique control effort minimizer of an $s$-integrator that passes $q$. Moreover, given with any function $\mathcal{K}(c,T)$, MINCO can also serve as a linear-complexity differentiable layer $\mathcal{W}(q,T):=\mathcal{K}(\mathcal{M}(q,T),T)$, such that $\partial\mathcal{W}/\partial{q}$ and $\partial\mathcal{W}/\partial{T}$ can be efficiently computed from any $\partial\mathcal{K}/\partial{c}$ and $\partial\mathcal{K}/\partial{T}$.

\begin{figure}[t]
    \centering
    \includegraphics[width=1.0\columnwidth]{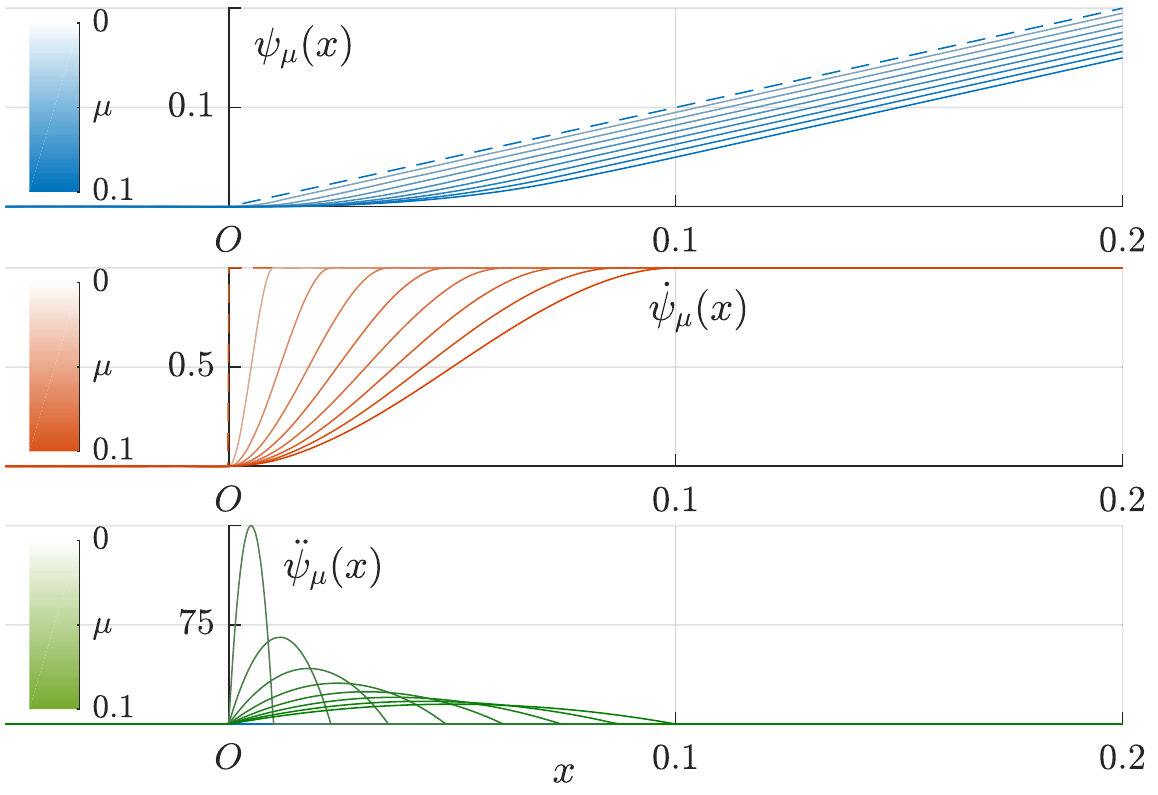}
    \caption{The function $\psi_\mu$ has continuous first and second derivatives for $\mu>0$. As $\mu$ approaches $0$, the function also approaches the exact penalty $\max(x,0)$. By iteratively shrinking $\mu$, constraints are enforced within any desired precision while keeping a bounded weight and the $C^2$-smoothness.\label{fig:SmoothedL1}}
    \vspace{-0.5cm}
\end{figure}

The corridor $\mathcal{S}$ provides natural constraints for the obstacle avoidance criterion in (\ref{eq:ObstacleAvoidanceCriterion}). We sequentially assign trajectory pieces into $\mathcal{S}$ as is shown in Figure~\ref{fig:TrajectoryInSFC}. For the case where a polyhedron $\mathcal{P}_i$ has only one piece, we have:
\begin{equation}
\label{eq:SpatialConstraints}
q_i\in\mathcal{P}_i\cap\mathcal{P}_{i+1},~\forall 1\leq{i}\leq{M},
\end{equation}
where $q_i$ is the $i$-th column in $q$. Unconstrained coordinates $q(\xi)$ and $T(\tau)$ are adopted such that (\ref{eq:SpatialConstraints}) and the positiveness of time are both satisfied by default~\cite{Wang2021Gcopter}. The continuous-time safety is enforced via the penalty functional below.
\begin{equation}
\label{eq:ObstacleAvoidanceFunctional}
\mathcal{I}_1(c,T)=\sum_{i=1}^M\int_0^{T_i}{\mathbf{1}\tp\phi_\mu\sBrac{A_ic_i\tp\beta(t)-b_i}\df{t}},
\end{equation}
where $\beta(t)=\rbrac{1,t,t^2,\dots,t^{2s-1}}\tp$ is the power basis, $\phi_\mu[\cdot]$ an entry-wise operator for $\phi_\mu$, and $\phi_\mu:\mathbb{R}\mapsto\mathbb{R}_{\geq0}$ an $C^2$-smoothing of the exact penalty. $\phi_\mu$ is defined as
\begin{equation}
\label{eq:C2SmoothingExactPenalty}
\phi_\mu(x)=
\begin{cases}
0 & \mathit{if}~x\leq0,\\
\rBrac{\mu-x/2}{\rBrac{x/\mu}^3} & \mathit{if}~0<x<\mu,\\
x-{\mu}/{2} & \mathit{if}~x\geq\mu.
\end{cases}
\end{equation}
Figure (\ref{fig:SmoothedL1}) shows that function $\phi_\mu$ approximates $\max(x,0)$ as $\mu$ approaches $0$, thus a finite weight for penalty can enforce the constraint at any specified precision.

The obstacle avoidance has been ensured by (\ref{eq:SpatialConstraints}) and (\ref{eq:ObstacleAvoidanceFunctional}), Now that they are defined via either $\cBrac{q,T}$ or $\cBrac{c,T}$, we can utilize the property of MINCO to formulate them on unconstrained coordinates $\xi$ and $\tau$. Note that there is no need for heuristic time allocation as done in traditional corridor-based methods. Optimization of decoupled time parameters are directly supported by MINCO.

\subsection{Robust Reciprocal Safety via Space-Time Capsules}

\begin{figure*}[t]
    \centering
    \includegraphics[width=1.9\columnwidth]{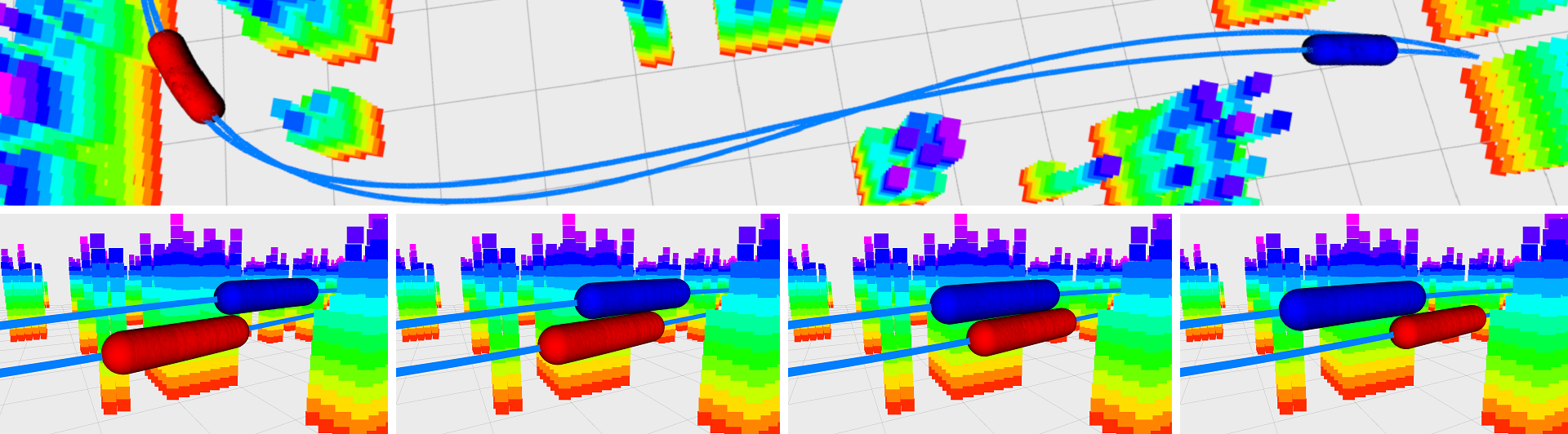}
    \caption{The top figure shows two vehicles fly in the opposite direction. The bottom figures show snapshots of their meeting. The red and blue capsules represent their spatial and temporal uncertainties. A robust planner should guarantee the condition (\ref{eq:ReciprocalSafetyCriterion}), i.e., no intersection occurs between them. In these figures, all trajectories are generated using the equivalent condition  (\ref{eq:EquivalentReciprocalSafetyCriterion}) instead. \label{fig:FlightInOppositeDirections}}
    \vspace{-0.5cm}
\end{figure*}

The reciprocal safety criterion (\ref{eq:ReciprocalSafetyCriterion}) ensures robustness in accidents that vehicles may fail to accurate trajectory tracking under wind disturbance or so on. Large errors can occur relative to the nominal position-stamp tuple. In this case, only considering the safe distance margin is insufficient because high-speed flights are common in large scenes.

The condition (\ref{eq:ReciprocalSafetyCriterion}) views vehicles as space-time capsules instead. Representing all uncertainties of a stamped position, a capsule is a sphere-swept volume of the trajectory segment on $[t-M_d,t+M_d]$ with sphere radius $M_r$. Robust reciprocal safety requires that all pairs of capsules be collision-free at any $t$. This condition is difficult to enforce since two time-varying nonconvex volumes are required to be collision-free all the time. Fortunately, an equivalent but more convenient condition exists:
\begin{subequations}
    \label{eq:EquivalentReciprocalSafetyCriterion}
    \begin{align}
    &\Norm{r(t)-r_i(\gamma)}_W\geq2M_r,\\
    &~\forall\gamma\in[t-2M_d,t+2M_d],\\
    &~\forall t\in[t_o,t_f].
    \end{align}
\end{subequations}

Here we give a proof of the equivalence between (\ref{eq:ReciprocalSafetyCriterion}) and (\ref{eq:EquivalentReciprocalSafetyCriterion}). For a given $r_i(t)$, if (\ref{eq:ReciprocalSafetyCriterion}) holds while there is a $\zeta\in[t_o,t_f]$ and a $\gamma\in[\zeta-2M_d,\zeta+2M_d]$ such that $\Norm{r(\zeta)-r_i(\gamma)}_W<2M_r$, then letting $\alpha=\zeta$, $\beta=\gamma$, and $t=(\zeta+\gamma)/2$ in (\ref{eq:ReciprocalSafetyCriterion}) gives a contradiction. If (\ref{eq:EquivalentReciprocalSafetyCriterion}) holds while there is an $\alpha\in[\zeta-M_d,\zeta+M_d]\cap[t_o,t_f]$ and a $\beta\in[\zeta-M_d,\zeta+M_d]$ such that $\Norm{r(\alpha)-r_i(\beta)}_W<2M_r$, then letting $t=\alpha$ and $\gamma=\beta$ in (\ref{eq:EquivalentReciprocalSafetyCriterion}) gives a contradiction.

It now becomes tractable to incorporate the robust reciprocal safety into our planning problem. Because (\ref{eq:EquivalentReciprocalSafetyCriterion}) only forbids collisions between a single volume and a dimensionless point at each timestamp. We enforce the space-time capsule constraint via the following penalty functional,
\begin{align}
\label{eq:CapsuleConstraintPenaltyFunctional}
&\mathcal{I}_2(c,T)=\\
&\sum_{i=1}^{N}\int_{t_o}^{t_f}\int_{-2M_d}^{2M_d}{\phi_\mu\rBrac{4M_r^2-\Norm{r(t)-r_i(t+v)}_W^2}\df{v}\df{t}},\nonumber
\end{align}
where $t_f$ and $r(\cdot)$ are only determined by $\cbrac{c,T}$. A planning result is given in Figure~\ref{fig:FlightInOppositeDirections}, showing the effectiveness of (\ref{eq:CapsuleConstraintPenaltyFunctional}).

\subsection{Physical Limits on Vehicle Dynamics with Drag Effects}
High-speed flights in large-scale scenes put forward higher requirements on dynamic feasibility. Unlike existing flatness-based methods using oversimplified dynamics, we enforce physical limits under nonlinear drag effects while still conducting optimization in the flat-output space.

Consider the vehicle state $x=\cBrac{r,\dot{r},R}$ where $r\in\mathbb{R}^3$ and $R\in\mathrm{SO}(3)$ are its translation and rotation, respectively. The input is $u=\cBrac{f,\omega}$ where $f\in\mathbb{R}_{\geq0}$ is the thrust and $\omega\in\mathbb{R}^3$ the body rates. The vehicle dynamics are defined as
\begin{subequations}
\label{eq:SymmetricDragDynamics}
\begin{empheq}[left=\empheqlbrace]{align}
    \label{eq:LinearAccelerationWithDrag}
    m\ddot{r}&=-mge_3-RDR\tp\sigma\rbrac{\norm{\dot{r}}}\dot{r}+Rfe_3,\\
    \dot{R}&=R\hat{\omega},
\end{empheq}
\end{subequations}
where $m$ is the vehicle mass, $g$ the gravitational acceleration, $e_3=(0,0,1)\tp$, $D=\diag\cbrac{d_h,d_h,d_v}$ a horizontally symmetric drag coefficient matrix, $\sigma:\mathbb{R}_{\geq0}\mapsto\mathbb{R}_{\geq0}$ a nonlinear term, and $\hat\omega$ a skew-symmetric matrix. Note we assume the vehicle to be horizontally symmetric. This decouples the yaw heading from drag effects and is common for multicopters as assumed in~\cite{Omari2013NonlinearCFD} and~\cite{Kai2017NonlinearFCFDE}. According to the analysis in~\cite{Bangura2012NonlinearDMHPC}, we adopt $\sigma(x)=1+C_{p}x$ in our lumped parameter model to incorporate both the linear drag~\cite{Faessler2018DiffFRD} and the parasitic drag.

The physical limits $\mathcal{G}(x,u)\preceq\mathbf{0}$ for (\ref{eq:SymmetricDragDynamics}) are defined as
\begin{equation}
\label{eq:PhysicalLimits}
\mathcal{G}(x,u)=\begin{pmatrix} \norm{\dot{r}}^2-v_{max}^2 \\ \norm{\omega}^2-\omega_{max}^2 \\ \arccos\rbrac{e_3\tp Re_3}-\theta_{max} \\ (f-f_{m})^2-f_r^2 \end{pmatrix}\preceq\mathbf{0},
\end{equation}
where $f_m=(f_{max}+f_{min})/2$ and $f_r=(f_{max}-f_{min})/2$ are intermediate constants. Maximum flight speed, body rate, tilt angle, and thrust are specified by $v_{max}$, $\omega_{max}$, $\theta_{max}$, and $f_{max}$, respectively. Besides, restricting the tilt angle prevents excessively aggressive maneuvers. Lower bounding the thrust by $f_{min}$ benefits attitude stabilization under disturbance.

The basic idea to incorporate physical limits on $x$ and $u$ into a flat trajectory is to utilize the algebraic transformation of differential flatness together with its differentiation. We denote by $r^{[s]}$ the stack of finite derivatives $\rbrac{r,\dot{r},\dots,r^{(s)}}$, and by $\psi$ the yaw. The flatness transformation is given by
\begin{equation}
\label{eq:FlatnessTransformation}
\rBrac{x,u}=\Psi\rbrac{r^{[s]},\psi^{[s]}}.
\end{equation}
All physical limits are enforced by the penalty functional,
\begin{equation}
\label{eq:PhysicalLimitFunctional}
\mathcal{I}_3(c,T)=\int_{t_o}^{t_f}{\mathbf{1}\tp\phi_\mu\sBrac{\mathcal{G}\circ\Psi\rBrac{r^{[s]}(t),\psi^{[s]}(t)}}\df{t}},
\end{equation}
where $\psi(\cdot)$ is any given planning of yaw. An energy functional is also incorporated to ensure a smooth flight,
\begin{equation}
\label{eq:TimeRegularizedSmoothness}
\mathcal{I}_0(c,T)=\int_{t_o}^{t_f}{\rBrac{\norm{r^{(s)}(t)}_2^2+\rho}\df{t}},
\end{equation}
where both $t_f$ and $r(\cdot)$ are still determined by $\cbrac{c,T}$.

Now we give all details about the algebraic function $\Psi$ for the concerned vehicle dynamics (\ref{eq:SymmetricDragDynamics}). We left multiply (\ref{eq:LinearAccelerationWithDrag}) by body axes $x_b=Re_1$ and $y_b=Re_2$, then
\begin{equation}
\rBrac{Re_i}\tp\rbrac{\ddot{r}+\frac{d_h}{m}\sigma\rbrac{\norm{\dot{r}}}\dot{r}+ge_3}=0,~\forall{i}\in\cBrac{1,2}.
\end{equation}
The consistency of $z_b=Re_3$ as $d_h$ vanishes implies
\begin{equation}
\label{eq:BodyAxisZ}
z_b=\mathcal{N}\rbrac{\ddot{r}+\frac{d_h}{m}\sigma\rbrac{\norm{\dot{r}}}\dot{r}+ge_3},
\end{equation}
where $\mathcal{N}(x)=x/\norm{x}_2$. Multiplying (\ref{eq:LinearAccelerationWithDrag}) by $z_b$ gives
\begin{equation}
f=z_b\tp\rBrac{m\ddot{r}+d_v\sigma\rbrac{\norm{\dot{r}}}\dot{r}+mge_3}.
\end{equation}
We use Hopf fibration~\cite{Watterson2019HOPF} to decompose the yaw quaternion $q_\psi$ and the tilt quaternion $q_z$ without involving Euler angles,
\begin{gather}
\label{eq:YawQuaternion}
q_\psi=\rBrac{\cos(\psi/2),0,0,\sin(\psi/2)}\tp,\\
\label{eq:TiltQuaternion}
q_z=\rBrac{1+z_b(3),-z_b(2),z_b(1),0)}\tp/\sqrt{2(1+z_b(3))}.
\end{gather}
Thus the rotation and body rates are given by
\begin{gather}
\label{eq:RotationMatrix}
R=\mathcal{R}_{quat}\rbrac{q_z\otimes{q_\psi}},\\
\label{eq:BodyRates}
\omega=2\rbrac{q_z\otimes{q_\psi}}^{-1}\otimes\rBrac{\dot{q}_z\otimes{q_\psi}+q_z\otimes\dot{q}_\psi},
\end{gather}
where $\otimes$ and the conversion $\mathcal{R}_{quat}(\cdot)$ are both given in~\cite{Vince2011quaternionsCG}, $\dot{q}_z$ and $\dot{q}_\psi$ are both evident from (\ref{eq:YawQuaternion}) and (\ref{eq:TiltQuaternion}), respectively.

Algebraic procedures (\ref{eq:BodyAxisZ})-(\ref{eq:BodyRates}) exactly define the function $\Psi$ in (\ref{eq:FlatnessTransformation}), whose gradient is as cheap as $\Psi$ itself~\cite{Griewank2008EvaluatingD}. Also, we know that $s=3$ is needed by the system (\ref{eq:SymmetricDragDynamics}) thus $r(t)$ should at least be jerk-controlled. There are only two intrinsic singularities in $\Psi$, which can be easily avoided by setting $\theta_{max}<\pi$ and $f_{min}>(d_v-d_h)\sigma\rbrac{v_{max}}v_{max}$. Our flatness map with nonlinear drag effects does not produce extra unnecessary singularities which occur in~\cite{Faessler2018DiffFRD}.

\subsection{Incremental Planning for Multi-Drone Coordination}
As for a newly occurred flight mission, we conduct the optimization below to accomplish trajectory planning subject to three concerned criteria in previous sections.
\begin{equation}
\label{eq:TrajectoryOptimization}
\min_{\xi,\tau}\sum_{\nu=0}^{3}w_\nu\cdot\mathcal{I}_\nu\rBrac{\mathcal{M}(q(\xi),T(\tau)),T(\tau)},
\end{equation}
where $w_0=1$, $w_1$, $w_2$, $w_3$ are all finite weights for (\ref{eq:C2SmoothingExactPenalty}), variables $\cbrac{\xi,\tau}$ are unconstrained coordinates for $\cbrac{q,T}$ in a fixed corridor. This unconstrained NLP can be solved efficiently and reliably by quasi-Newton methods. Due to the fact that optimization usually focuses on high-quality local solutions, the initial values for points $q$ and the time allocation $T$ need to be further specified.

\begin{figure}[ht]
    \centering
    \includegraphics[width=0.95\columnwidth]{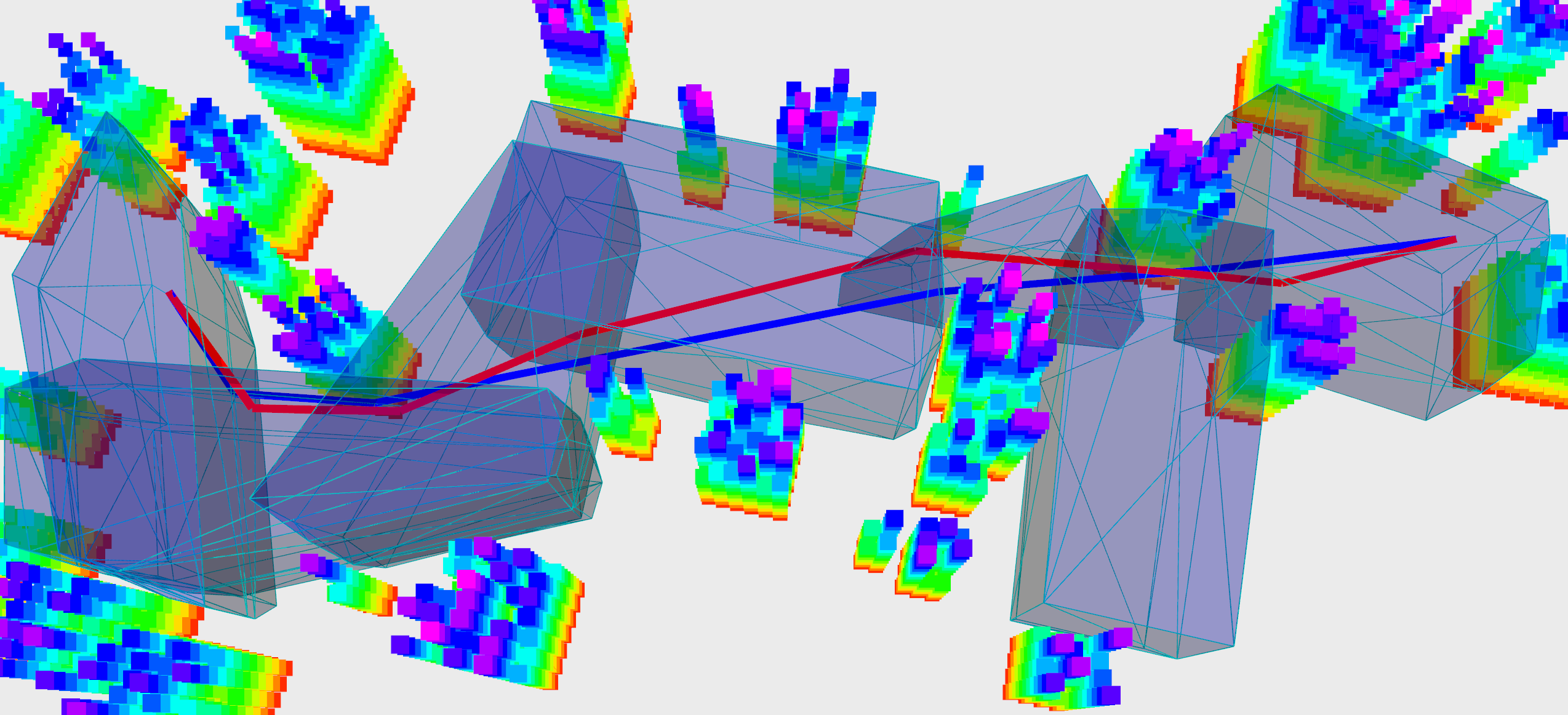}
    \caption{The red path is generated by Informed RRT*~\cite{Gammell2018InformedRRT}. The blue path is generated by distance-minimization (\ref{eq:ShortestPathInCorridor}) with $\sigma=10^{-2}$. The blue one is a high-quality homotopic refinement computed within $\SI{0.025}{\milli\second}$.\label{fig:ShortestPath}}
    \vspace{-0.5cm}
\end{figure}

As for $q$, we take the waypoints from a shortest path as its initial value. Denote $x_0=p_o$ and $x_M=p_f$. The path is given by a lightweight distance minimization, i.e.,
\begin{subequations}
    \label{eq:ShortestPathInCorridor}
    \begin{align}
    \min_{\xi}&\sum_{i=1}^M\sqrt{\norm{x_i-x_{i-1}}^2_2+\delta},\\
    ~s.t.&~x_i=q_i(\xi_i),~1\leq{i}<M.
    \end{align}
\end{subequations}
Here we use $\sqrt{\norm{\cdot}^2_2+\delta}$ as a $\delta$-smooth approximation~\cite{Beck2012SmoothingAUF} to the Euclidean distance, where $\delta$ takes the desired precision. Note that coordinates $\xi$ preserve local minima of the original convex problem for $q$. This minimization efficiently refines the rough solution as depicted in Figure~\ref{fig:ShortestPath}.

As for $T$, we set the initial guess using trapezoidal velocity profiles~\cite{Lynch2017ModernROB} for the obtained $q$. A $p(t):[0,\bar{T}]\mapsto\mathbb{R}^3$ is then generated by (\ref{eq:TrajectoryOptimization}) with $w_2=0$, i.e., reciprocal safety is not considered. If $p(t)$ happens to be safe by checking (\ref{eq:EquivalentReciprocalSafetyCriterion}), the planning is completed. Otherwise, temporal scheduling should be done for the length-parameterized curve of $p(t)$. We conduct kinodynamic RRT*~\cite{Karaman2010OptimalKDMP} for a one-dimensional double integrator along the curve length, with limited speed and acceleration. Its cost and steering function are all determined by a $1$-dimensional time-optimal curve whose closed-form solution is trivial for a double integrator. Note that the time scheduling never fails since a feasible solution always exists if the vehicle waits for a long enough time. The result provides a feasible time allocation $T$. Along with waypoints of $p(t)$, we finally obtain an initial feasible guess for (\ref{eq:TrajectoryOptimization}).

\section{Results}
\subsection{Implementation Details}
\begin{figure*}[t]
    \begin{center}
        \subfigure[\label{fig:LargeMapMultiDronePlanning} Multi-drone planning for $80$ and $160$ vehicles in a $\SI{4}{\square\kilo\meter}$ dense-obstacle environment. Two left figures use distance-based safety constraints. Two right figrues use space-time capsule constraints instead. All trajectories fulfill the physical limits under drag effects.]
        {\includegraphics[width=1.9\columnwidth]{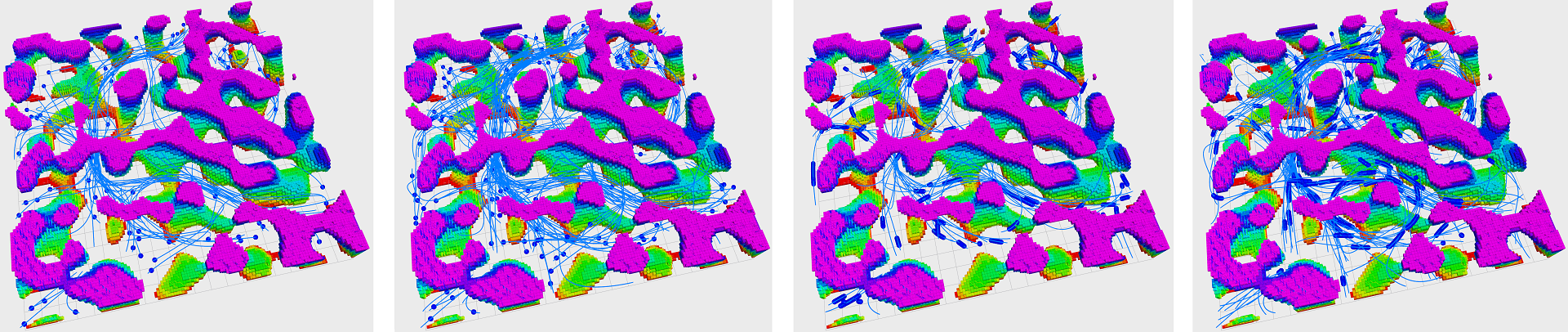}}
        \subfigure[\label{fig:ConstraintsFlightDrag} The dynamic profile of a vehicle in the fourth figrue above. Dashed lines indicate physical limits. The mass normalized thrust, speed, tilt angle, and the magnitude of body rate are all provided. The mass-normalized magnitude of the drag force is also given in the light-blue curve.]
        {\includegraphics[width=1.9\columnwidth]{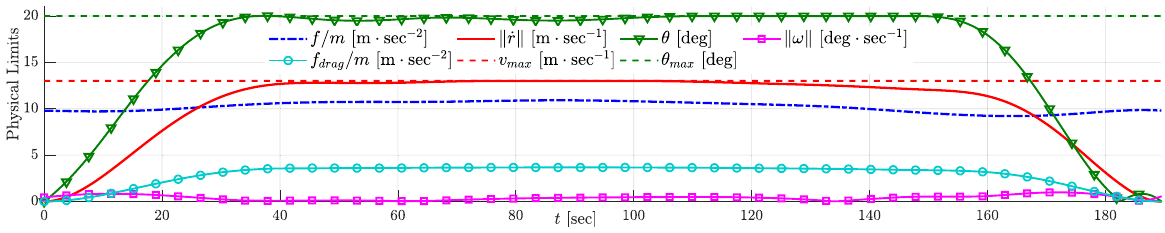}}
    \end{center}
    \vspace{-0.3cm}
    \caption{\label{fig:LargeMapMultiDronePlanningProfile} Two subfigures give both the planning results in a large scene and the dynamic profile of a selected vehicle.}
    \vspace{-0.5cm}
\end{figure*}
We conduct multi-drone trajectory planning in an $2000\times2000\times200m^3$ space where obstacles are randomly generated by~\cite{Wu2019Mockamap}. The occupancy map is firstly polyhedronized with $\epsilon=10^{-5}$, implying the filling rate to be $99.999\%$. In safety criteria, we use $M_d=\SI{4}{\second}$ and $M_r=\SI{15}{\meter}$. In vehicle dynamics, we use $m=\SI{1.9}{\kilo\gram}$, $d_h=d_v=\SI{0.475}{\per\second}$, $C_p=\SI{0.01}{\per\meter\second}$. We set physical limits as $v_{max}=\SI{13}{\meter\per\second}$, $\omega_{max}=\SI{2\pi/3}{\radian\per\second}$, $\theta_{max}=\SI{\pi/9}{\radian}$, $f_{min}=\SI{9.5}{\newton}$, and $f_{max}=\SI{28.5}{\newton}$. This implies the thrust-to-weight ratio to be only $1.53$. Besides, $\mathcal{I}_1$, $\mathcal{I}_2$, and $\mathcal{I}_3$ are evaluated via equally-spaced quadrature with fixed node numbers. In optimization, we use $\rho=10^{-3}$ for time regularization, $\mu=10^{-2}$ as the initial smooth factor, and $w_1=w_2=w_3=10^{5}$ as weights, which ensure a high precision for constraints. Based on these settings, we plan for every newly-generated pair of start and goal in the map.

\subsection{Multi-Drone Planning - Physical Limits}
We enforce both the distance-based constraints and our space-time capsule constraints in (\ref{eq:ReciprocalSafetyCriterion}). The results are shown in Figure~\ref{fig:LargeMapMultiDronePlanningProfile}. All trajectories are generated in several minutes. In Figure~\ref{fig:LargeMapMultiDronePlanning}, our scheme ensures the safety of concurrent flights for up to $160$ vehicles. Moreover, physical limits under drag effects are considered even if most vehicles have to cover about a kilometer. We give profiles of physical limits in Figure~\ref{fig:ConstraintsFlightDrag} for one of the vehicles, implying that all constraints (\ref{eq:PhysicalLimits}) are satisfied during the entire $\SI{3}{\minute}$ flight.

According to Figure~\ref{fig:ConstraintsFlightDrag}, our planner differs from traditional ones in that it does not assume a point-mass model. For example, some multi-drone delivery missions require a maximum tilt angle of fragile payloads. In this case, our planning directly meets these requirements as the actual tilt angle $\theta$ is always below $\theta_{max}$. Moreover, it outperforms traditional ones that assume a drag-free rigid body. For example, whenever a vehicle flies as a large constant velocity, its acceleration becomes zero, and its thrust is exactly the weight if no drag is considered. Actually, the vehicle produces more thrust $f$ and a nonzero tilt angle $\theta$ to cancel drag forces $f_{drag}$ as shown in our figure. Therefore, multi-drone flights in large schemes indeed involve more realistic factors.

\subsection{Multi-Drone Planning - Robustness}
\begin{figure}[ht]
    \centering
    \includegraphics[width=1.0\columnwidth]{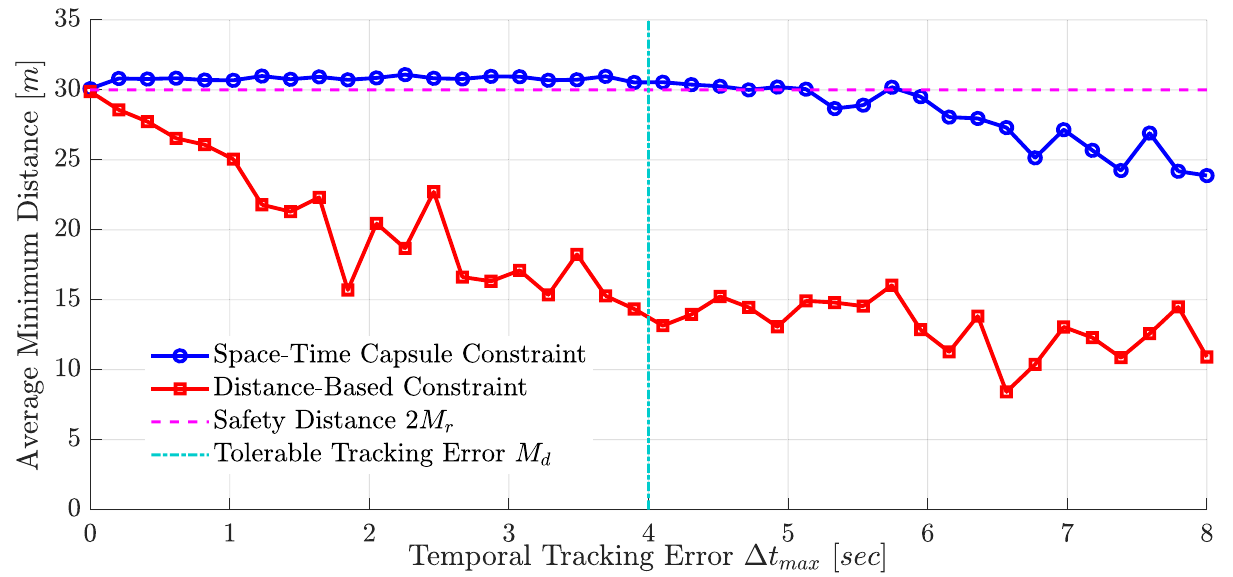}
    \caption{Average minimum distances among all vehicle pairs for different safety criteria and temporal tracking error.\label{fig:RobustnessTestResults}}
    \vspace{-0.5cm}
\end{figure}

To validate the robustness of our scheme. We consider a case where $20$ vehicles concurrently fly through a narrow gap, as shown in Figure~\ref{fig:RobustnessTestViz}. We use both the distance-based safety constraints and our capsule constraints to solve this planning. Moreover, we add disturbance to the vehicle such that tracking error occurs during their high-speed flight. Denote by $\Delta{t}_{max}$ the maximum temporal error for nominal trajectory tracking. For different $\Delta{t}_{max}$, we compute the minimum distance among all vehicle pairs. The entire multi-drone planning is repeated for $20$ times. The average minimum distance is counted for both constraints in Figure~\ref{fig:RobustnessTestResults}. As $\Delta{t}_{max}$ becomes positive, distance-based safety is quickly broken since the red curve goes below the safe distance $2M_r=\SI{30}{\meter}$. Our capsule constraints guarantees the safety when $\Delta{t}_{max}\leq{M_d}$. Moreover, the safe distance is still maintained even if $\SI{4}{\second}<\Delta{t}_{max}<\SI{5}{\second}$.

\section{Conclusion}
In this paper, we propose a systematic scheme for robust multi-drone planning at high speeds. The free-space-oriented map much eases our planner from the burden of high-volume data accessing. The space-time capsule constraint ensures reciprocal safety even if any vehicle is significantly behind the predefined flight progress. The minimum-singularity flatness of our drone dynamics subject to nonlinear drags plays an essential role in ensuring realistic physical limits at high speeds. We believe this is a practical framework towards robust multi-drone trajectory planning in large scenes.

\bibliography{references}

\end{document}